\documentclass{article}
\usepackage{preamble}

\title{ Explaining CLIP through Co-Creative Drawings and Interaction
\vspace{-0.1cm}
}

\author{
Varvara Guljajeva*‡\textsuperscript{1}, 
Mar Canet Solà*‡\textsuperscript{2},  
Isaac Joseph Clarke\textsuperscript{1}\\
\small \textsuperscript{1}The Hong Kong University of Science and Technology. Guangzhou, China\\
\small \textsuperscript{2}Baltic Film, Media and Arts School, Tallinn University\\  
\small ‡Corresponding authors: varvarag@ust.hk, mar.canet@tlu.ee \\
\small *equal contribution as first authors  
}

\date{\small June 12, 2023 \vspace{-0.4cm}} %
\begin{document}
\maketitle

\begin{center}
    \centering
    \includegraphics[width=\textwidth]{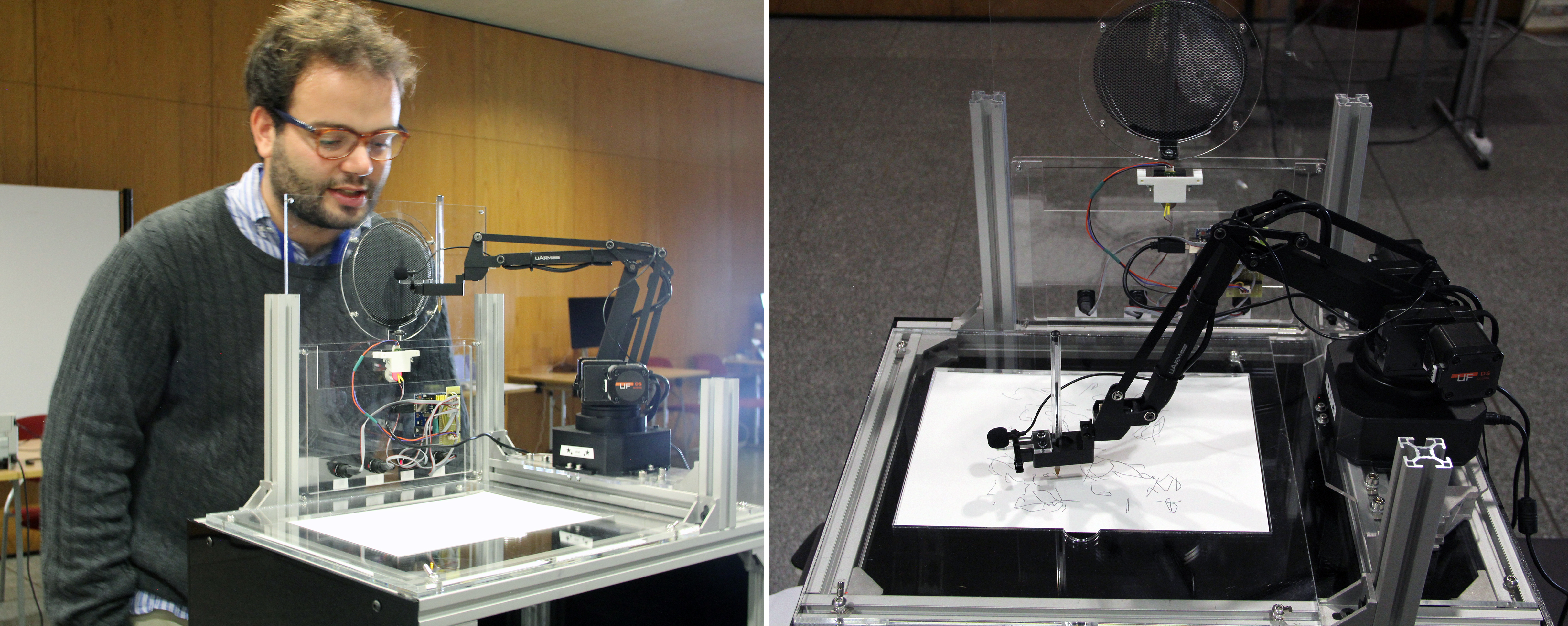}
    \captionof{figure}{Dream Painter installation at ACM Multimedia 2022 Conference. On the left: a participant interacting with the installation by telling a dream to the robot. On the right: the robot drawing CLIP-generated line drawing from the speech input.}
    \label{fig: small_dreampainter}
\end{center}

\begin{abstract}
This paper analyses a visual archive of drawings produced by an interactive robotic art installation where audience members narrated their dreams into a system powered by CLIPdraw deep learning (DL) model that interpreted and transformed their dreams into images. The resulting archive of prompt-image pairs were examined and clustered based on concept representation accuracy. As a result of the analysis, the paper proposes four groupings for describing and explaining CLIP-generated results: clear concept, text-to-text as image, indeterminacy and confusion, and lost in translation. This article offers a glimpse into a collection of dreams interpreted, mediated and given form by Artificial Intelligence (AI), showcasing oftentimes unexpected, visually compelling or, indeed, the dream-like output of the system, with the emphasis on processes and results of translations between languages, sign-systems and various modules of the installation. In the end, the paper argues that proposed clusters support better understanding of the neural model.
\end{abstract}

\begin{figure*}[!t]
    \centering
    \includegraphics[width=1\textwidth]{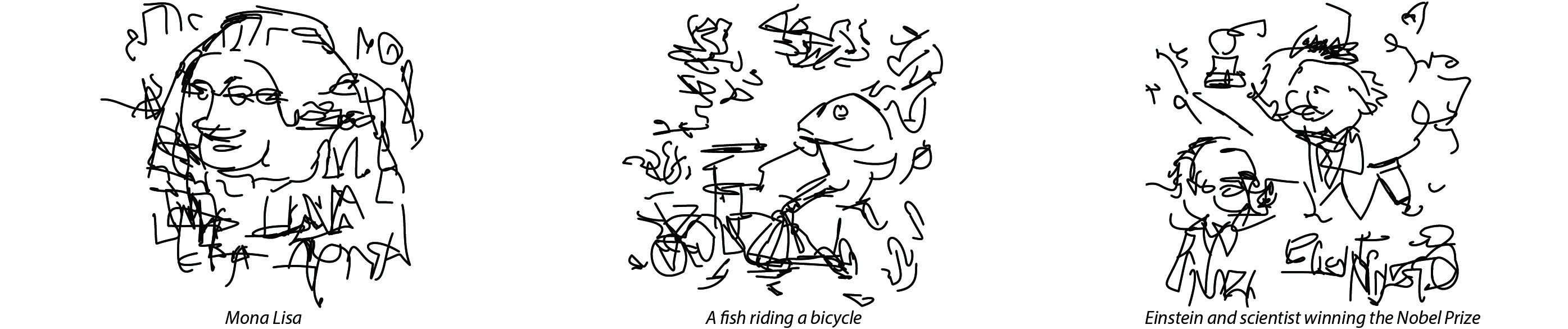}
    \captionof{figure}{An example of grouping 1: Clear Concepts}
    \label{fig: Clear Concepts}
\end{figure*}

\section{Introduction}
Often AI is referred to as ‘a black box’. Complex technical descriptions given to explain neural networks create more confusion than clarity for an average person. Explainable AI aims to increase the transparency of AI systems and our understanding of the decisions of AI algorithms. Generative models produce artefacts rather than decisions or forecasts, and it is necessary to explore the construction of these outputs and their origins in other ways  \parencite{sun2022investigating}. Experiential and interactive applications of these models can aid our exploration of the limitations and biases of these models by making the outputs tangible to a wider audience, where the mechanisms can be negotiated collaboratively.
\newline
\indent Artists have been deploying AI and robotics in drawing. One example is AARON by Harold Cohen which originates from the early 1970s \parencite{cohen2016harold}. Modern creative AI continues to expand artists' toolsets, possibilities for novel art forms, and cross-disciplinary connections. One such DL tool is the neural network CLIP, released by OpenAI in 2021 and trained on image and text pairs \parencite{radford2021learning}. This model represents images and texts as 512-number vectors. This shared space allows text-image comparisons. We can encode an image and multiple text descriptions, then compare the distances between the encodings to see which text labels best represent the image content. The CLIPdraw algorithm repeatedly adjusts a random arrangement of lines, to move the image embedding closer to the text prompt embedding. This process of guided adjustments allows us to translate a text prompt into an image. CLIP guidance has been widely adopted in text-to-image models to guide GANs and diffusion processes.
Image generation with CLIP is limited by the data it has been trained on. The original CLIP paper notes a 400 million image-text pair dataset \parencite{radford2021learning}. We do not know what images and texts were in this dataset, but by examining the drawings generated we can speculate on the contents.\newline
\indent In this paper, we use audience interaction and experience to explain how CLIP works and witness its limitations. The drawings presented here originate from the interactive robotic art installation Dream Painter by Varvara \& Mar, which was a part of the Art Gallery at ACM Multimedia 2022. Through the interactive experience of speech-to-image translation, a user can navigate in the latent space of a DL model called CLIP, with the algorithm CLIPdraw \parencite{frans2022clipdraw}, which results in an image  drawn by a robot  \parencite{guljajeva2022dream,canet2022dream}. This approach distinguishes itself from pixel-based text-to-image models, such as DALL-E, Midjourney, and Stable Diffusion. It provides a distinct audience experience by sketching the dreams and creating visually open and interpretive outputs.
Due to the time limit set by the interactive real-time system, the algorithm runs 100 steps trying to converge the lines to text in 15 seconds. The original-sized installation uses an industrial Kuka arm robot with a multicolored painting system. The images presented here originate from a small version of the artwork that uses a single color and a smaller uArm robot. The audience shares their dreams by talking into a microphone, their words then guide the image generation process, and the robotic arm draws a picture representing their dream onto A4 paper (see Figure \ref{fig: small_dreampainter}).

\section{Classification}
In terms methodology applied, we present groupings of drawings, through which we initiate a discussion regarding intersemiotic translatability of concepts and, ultimately, the explainability of AI. The visual analysis was performed by four researchers taking into account the audience's observations and informal discussion with them. Prompt-image pairs constitute the bulk of the visual content, representing the system's input and output  and documenting the interactions during the exhibition. A close reading of the collected drawing was then conducted. The fifty-one drawings produced were organised into four groups that reveal different behaviours of CLIP: the drawings that demonstrated the concept of user input clearly, the drawings that output drawn text instead of figures, the drawings that partly contained the concept of the input, and the drawings that did not match the concept of the dream.

\subsection{Clear Concepts}
The first group features clear concepts where the content of the drawing is understandable, and the prompt can be guessed. Informal discussion with 51 participants showed that the images with clear concept prompts behind them were the most easily guessed. Objects and the relations between them are relatively clear, with straightforward, short prompts resulting in minimal mistranslations. 
This group of images demonstrate the model's capacity to translate dream prompts into expected images. At a certain level, the process of translation functions as we would expect, familiar concepts result in familiar images. Dreams are often uncertain, with unfamiliar concepts and jarring relationships between objects. Knowing the baseline at which the model responds as expected helps us understand where and how it fails. Understanding failure in deep learning models can, in turn, help explain the internal representations these models have of the world, and can also teach us how to use these tools in creative pursuits. The Mona Lisa drawing serves as a reliable waypoint or an “island of sense” in our navigation of CLIP’s latent space \parencite{nancy2013pleasure}.

There are a few interesting elements to \textit{Mona Lisa} that we observe. The robot generates a drawing that not only resembles the iconic face, but also includes text scrawled around the image (see Figure \ref{fig: Clear Concepts}). We can see several Ms and Ls. Speculating on content included in the dataset used to train the model, it appears that \textit{Mona Lisa} has been connected to images other than the original portrait; posters, merchandise, photography, or other reinterpretations. Similar qualities can be seen in the drawing of Einstein.

\begin{figure*}[!t]
    \centering
    \includegraphics[width=\textwidth]{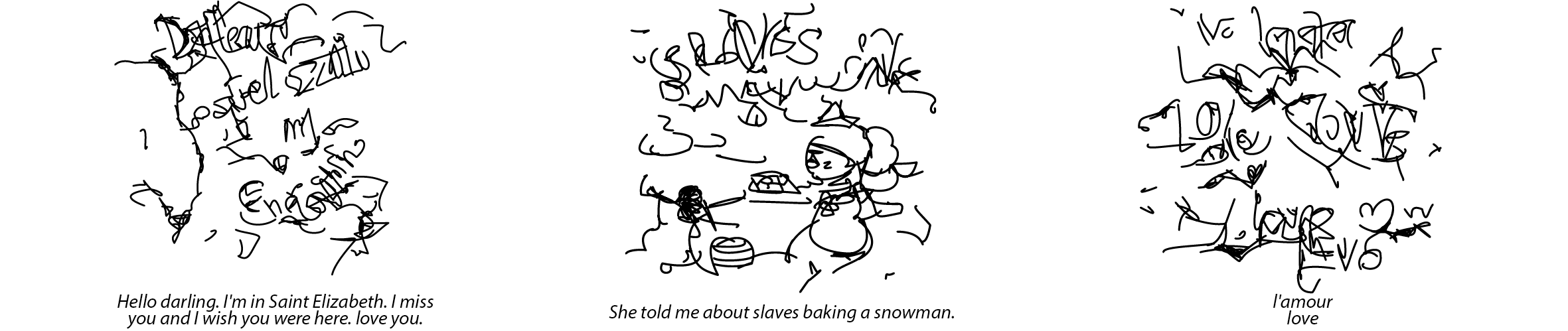}
    \caption{An example of grouping 2: Text-to-text as Image}
    \label{fig: text-as-image}
\end{figure*}

\subsection{Text-to-text as Image}
In the second grouping of images we have identified instances where the text prompt has been drawn into a text-image. These text-images show the connections words have in the model. The drawing has been guided towards writing words that don’t appear in the prompt but are related, for example, the drawing prompt \textit{L’amour} seems to be made up of many copies of the word Love (see Figure \ref{fig: text-as-image}). Our restriction of single-color drawing may also be biasing the algorithm towards certain outputs A black heart would give a very different reading to a red heart, instead, it is being drawn towards textual representation. Text-dominant drawings  also relate to how we place text in an image; the design of posters, user interfaces, and calligraphy. In the introduction, we discussed how training data influences the types of images that can be drawn. When we examine this grouping of images we question if the image of text is the best representation, or used due to limits of the training data.\newline
\indent In the drawing \textit{Hello darling I’m in Saint Elizabeth I miss you and I wish you were here love you} we see a different kind of prompt given that goes beyond the artist's request for the audience to share their dreams. Instead, the audience member has used the artwork as a way to transmit a message to a loved one. The drawing resembles the writing seen on gift cards; large imitation hand-drawn lettering centred in the image, with frilly decoration surrounding the text. The love letter prompt has guided the drawing towards a commonly known Valentine's Day card design, again demonstrating how text, images, and images of text, all occupy a shared space in the model. 

The influence of the initial state, the random seed, and other constraints like colour palette, is revealed. The frequent occurrence of text-images should be expected when starting with noisy black lines on a white background.

\begin{figure*}[!t]
    \centering
    \includegraphics[width=\textwidth]{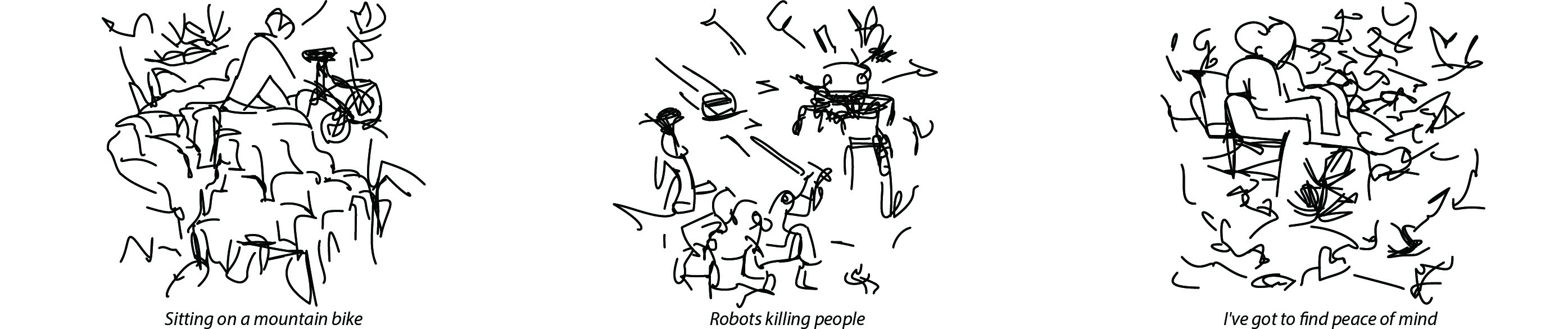}
    \caption{An example of grouping 3: Indeterminacy And Confusion}
    \label{fig: indeterminacy}
\end{figure*}

\subsection{Indeterminacy And Confusion.}
In the first group of images the concepts are clear and the combination of ideas is easy for us to picture in our minds, then in this grouping CLIP understood only partly the concept and failed to depict the meaning.

Despite this large number of training examples in the CLIP dataset, it is easy for us to imagine arrangements of objects and ideas that have never been seen, particularly when thinking about our dreams where rules of physics, or the usual behaviours of objects do not apply. CLIP may have seen many images of cats wearing hats, but it is unlikely to have seen a hat wearing a cat. CLIP struggles with guiding unusual arrangements of concepts. In the drawing \textit{Robots Killing People}, we see what appears to be people killing robots (see Figure \ref{fig: indeterminacy}). CLIP appears to have understood Robots, Killing, and People, as elements to be included but we end up with a drawing quite the opposite in meaning. 

In \textit{Sitting on a mountain bike} we see a loose drawing of a character sitting on a mountain, with a bike sticking out, as though it is a misplaced object, it is as though it has drawn \textit{Sitting on a mountain} and then appended \textit{bike} as a separate element. Again, we see that concepts are known by CLIP, but the relationships fall apart and the meaning is lost. It is important to be aware when being guided by these models that they reflect the patterns and associations in the datasets they are trained on, and there are limitations in attempting to deviate from expected compositions.

\subsection{Lost In Translation}
\begin{figure*}[!t]
    \centering
    \includegraphics[width=\textwidth]{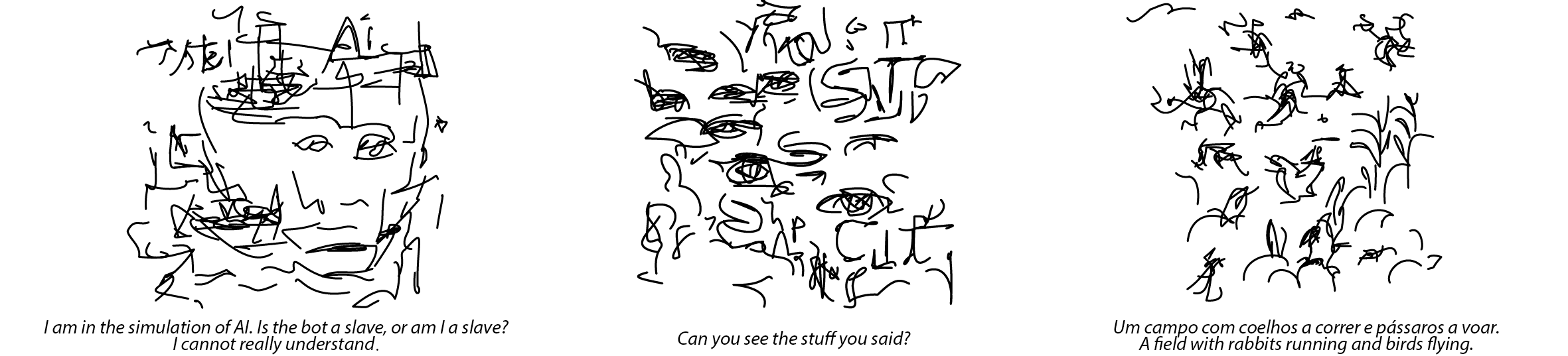}
    \caption{An example of grouping 4: Lost In Translation}
    \label{fig: lost}
\end{figure*}

With this group of drawings, unlike \textit{Mona Lisa} or \textit{A fish riding a bicycle}, it is difficult to guess what the prompt would be from seeing the drawing. They are visually interesting, but hard to deconstruct. In some cases this ambiguity may be due to equally uncertain prompts, in others, we find after reading the prompt we begin to see what has been drawn. For example, in \textit{Can you see the stuff you said?} we can see shapes of eyes hidden in the noisy scribbles, shapes that may be unclear without first being aware of the prompt (Figure\ref{fig: lost}).

Aaron Hertzman has described how GAN art has a quality of visual indeterminacy, where elements of the image seem coherent but on closer examination confound explanation \parencite{hertzmann2020visual}. He attributes this lack of stability in artworks as a consequence of “powerful-but-imperfect image synthesis” models. These drawings, although vector-based line drawings, not full-color pixel images, display a similar quality of indeterminacy. 

\textit{I am in the simulacrum of AI the boat is a slave or I’m a slave of the but I cannot really understand} is a prompt full of uncertainty and ambiguity. Dreams are often hard to remember, made of conflicting ideas and unresolved stories. Whilst recalling their dream, the dreamer realizes they aren’t quite sure what happened, and this uncertainty permeates the many layers of translation leading to the eventual drawing. In this example, the initial mistranslation from speech-to-text had a large effect on the confusion in the prompt. The participant had said the word ‘bot’, as in robot, and this was recorded as boat. What began as a comment on AI turned into a more dreamlike image when processed through [Artwork Anonymised]. The drawing is guided towards faces \textit{(I am)}, boats and waves \textit{(boat / slave)}, and combines these with unclear lettering \textit{(I cannot really understand)}.

\section{Discussion}
We have outlined a few overlapping clusters that show the variety of images that can be generated by CLIP guidance. Although the prompts submitted to the system were more spontaneous than engineered, due to the real-time nature of the art installation, this imperfection in prompts triggered unexpected creativity and understanding of the algorithm's logic. According to Juri Lotman, illegitimate imperfections create new and unexpected possibilities of meaning that result in creativity \parencite{lotman1990universe}. 

Firstly, engaging with the interactive robotic installation provided a novel experience for the audience. On average, they spent 10 minutes with the artwork, interacting, observing the drawing process, and subsequently analyzing and discussing the output as a paper drawing. We surveyed 51 participants, asking them how representative the picture was of their dream on a scale of ten. The average score obtained was 6.7. This indicates that most people comprehended what was depicted in the drawing and how CLIP represented certain elements. The audience awarded fewer points when they noticed contextual inaccuracies, such as a mountain bike sticking out of the hill rather than riding on top of the mountain. On the other hand, the imperfections of CLIP made the audience laugh and the experience with the project enjoyable. We believe a physical and multimodal interface made the audience spend more time with the installation and analyse the paper drawing afterwards, which also contributed towards understanding how text-to-image model works. 

What is evident in this process is that the quality of the prompt is critical to the quality of the drawing returned. Several papers on audience interaction with AI-aided artworks emphasise the importance of the human part in valuable output generation on the AI side \parencite{canet2022dream,guljajeva2021synthetic,guljajeva2022dream}. Here we are referring to meaningful interaction and not prompt engineering. It might be that some more complex concepts that are classified in 3 and 4 categories could result in closer to the prompt drawings by running more steps in the algorithm. However, in the case of this study, it was less important than audience's experience while interacting with the installation.

Prompt engineering is critical to controlling the output of text-to-image generation. Wittgenstein, in their philosophical proposition in the Tractatus, explores the connection between the notions of "What can be shown cannot be said" and "Whereof one cannot speak, thereof one must be silent." \parencite{wittgenstein1999tractatus}. These concepts shed light on the inherent limitations of language when try to describe an image and the communicative affordances of visual imagery vs language. Moreover, we cannot refine or edit our prompt when interacting with this artwork. We are restricted to the order of words as they leave our mouths at the moment of interaction. An audience member may approach the work slightly nervous, lacking precision with their choice of language. Someone more familiar with this technology may deliberately alter their speech to be clearer for a machine. By adding extra boundaries of translation, we remove the possibility of overthinking and overanalyzing the input, the audience hands over a loose dream, placing trust in the chance operations of the system.\newline
\indent We also translate the spoken language. The audience could choose between English, Spanish, Portuguese, or French. Each translation process adds extra noise into the system. Dream Painter takes chance arrangements and imprecise translations to explore order and disorder in AI models. The drawings included in this paper highlight technical and communicative acts of translation between different subsystems of the work. By probing the thresholds and boundaries between distinct semiotic spaces within a heterogenous semiosphere of the work we address the questions of limits of intersemiotic translation or, in Roman Jakobson’s words, “transmutation” \parencite{jakobson2002linguistic} between distinct elements or subdomains of complex technical systems, and tension between the ethical ideal of explainable and transparent AI and mystery and ambiguity often attributed to the work of art.

We can learn how generative AI models work by interacting with them. By clustering and examining these drawings, we can understand how changes to the prompt can drastically alter the images, and can see how certain uses of language, in combination with representational constraints, can teach us how to guide these processes.

\section{Conclusion}
This paper presents our interpretation and grouping  of AI-generated drawings in response to dreams shared by the audience. These drawings show how the responses of generative AI algorithms are heavily determined by both the quality of the user input and the content of the dataset the models were trained on. This work demonstrates how meaning can be distorted through layers of translation, from speech-to-text, to vector encodings, to physical drawing, and how uncertainty can permeate these boundaries of technology. At the same time, imprecision and mistranslation of input led to unexpected results that contributed to creativity and the discovery of the logic behind the technology. The novel interaction experience with the robot and CLIP model made people spend time with the installation and analyse their experience and result. Thus, we believe that by experiencing the translation process through a physical and artistic interface has a positive effect on understanding how DL models make such translations, and on creativity that results from unexpected interaction results with the system. 

The clusters we have identified show how well-known imagery has a clear presence in the model. Still, the inability to handle unusual arrangements can cause drawings to have drastically different readings from the original prompt. We have seen how some concepts are drawn as images of texts, in some cases because of hard-to-visualise words, and in other cases, the constraints of the drawing favouring textural representation. With Dream Painter, we have shown how interesting and unexpected drawings can emerge due to CLIP guidance. 

\section{Author contributions}
VG and MCS are the authors of artistic idea and realisation of Dream Painter project. VG and MCS collected and analysed the drawings, surveyed the audience, and write the paper. IJC participated in analysing the drawings and writing the article.

\section{Acknowledgments}
MSC is funded through the EU Horizon 2020 research and innovation program (Grant No.810961). Thanks to Yue Huang for designing Figures 2-5, and to Iurii Kuzmin for participating in the initial drawings' analysis discussion. 
 
\begingroup
\setlength{\emergencystretch}{8em}
\printbibliography
\FloatBarrier
\endgroup

\end{document}